\documentclass[letterpaper]{article}
\usepackage{aaai20}
\usepackage{times}
\usepackage{helvet}
\usepackage{courier}
\usepackage[hyphens]{url}
\usepackage{graphicx}
\usepackage{pdfpages}

\usepackage{comment}
\usepackage{wrapfig}
\usepackage{subcaption}
\usepackage{enumitem}
\usepackage{bbm}
\usepackage{amsmath}

\newcommand{\citet}[1]{\citeauthor{#1} \shortcite{#1}} \newcommand{\citep}{\cite}

\title{Navigating the Trade-Off between Multi-Task Learning and Learning to Multitask in Deep Neural Networks}

\author{\\{\bf Sachin Ravi$^{1}$, Sebastian Musslick$^{2}$, Maia Hamin$^{1,2}$, Theodore L. Willke$^3$ and Jonathan D. Cohen$^{2,4}$}\\ 
        $^1$Department of Computer Science, Princeton University, Princeton, NJ 08544, USA.\\
        $^2$Princeton Neuroscience Institute, Princeton University, Princeton, NJ 08544, USA.\\ 
        $^3$Parallel Computing Lab, Intel Corporation, Santa Clara, CA 95054, USA. \\
        $^4$Department of Psychology, Princeton University, Princeton, NJ 08544, USA.\\ 
        $^*$Corresponding Author: {\it sachinravi14@gmail.com}} 

\begin{document}
\raggedbottom

\maketitle

\begin{abstract}
The terms multi-task learning and multitasking are easily confused. Multi-task learning refers to a paradigm in machine learning in which a network is trained on various related tasks to facilitate the acquisition of tasks. In contrast, multitasking is used to indicate, especially in the cognitive science literature, the ability to execute multiple tasks simultaneously. While multi-task learning exploits the discovery of common structure between tasks in the form of shared representations, multitasking is promoted by separating representations between tasks to avoid processing interference. Here, we build on previous work involving shallow networks and simple task settings suggesting that there is a trade-off between multi-task learning and multitasking, mediated by the use of shared versus separated representations. We show that the same tension arises in deep networks and discuss a meta-learning algorithm for an agent to manage this trade-off in an unfamiliar environment. We display through different experiments that the agent is able to successfully optimize its training strategy as a function of the environment.

\end{abstract}

\section{Introduction}

Many recent advances in machine learning can be attributed to the ability of neural networks to learn and to process complex representations by simultaneously taking into account a large number of interrelated and interacting constraints - a property often referred to as parallel distributed processing \citep{mcclelland1986appeal}. Here, we refer to this sort of parallel processing as interactive parallelism. This type of parallelism stands in contrast to the ability of a network architecture to carry out multiple processes independently at the same time. We refer to this as independent parallelism and it is heavily used, for example, in computing clusters to distribute independent units of computation in order to minimize compute time. Most applications of neural networks have exploited the benefits of interactive parallelism \citep{bengio2013representation}. For instance, in the multi-task learning paradigm, learning of a task is facilitated by training a network on various related tasks \citep{caruana1997multitask,collobert2008unified,kaiser2017one,kendall2018multi}. This learning benefit has been hypothesized to arise due to the development of shared representation between tasks \citep{baxter1995learning,caruana1997multitask}. However, the capacity of such networks to execute multiple tasks simultaneously\footnote{Here we refer to the simultaneous execution of multiple tasks in a single feed-forward pass.} (what we call multitasking) has been less explored.

Recent work \citep{musslick2016controlled,musslick2017multitasking,PetriInPrep} has hypothesized that the trade-off between these two types of computation is critical to certain aspects of human cognition. Specifically, though interactive parallelism allows for quicker learning and greater generalization via the use of shared representations, it poses the risk of cross-talk, thus limiting the number of tasks that can be executed at the same time (i.e. multitasking). Navigation of this trade-off by the human brain may explain why we are able to multitask some tasks in daily life (such as talking while walking) but not others (for example, doing two mental arithmetic problems at the same time). \citet{musslick2017multitasking} have shown that this trade-off is also faced by artificial neural networks when trained to perform simple synthetic tasks. This previous work demonstrates both computationally and analytically that the improvement in learning speed through the use of shared representation comes at the cost of limitations in concurrent multitasking \cite{musslick2020rationalizing}.

While these studies were informative, they were limited to shallow networks and simple tasks.\footnote{See \citet{Alon2017,musslick2020rational} for a graph-theoretic analysis of multitasking capability as a function of network depth.} Moreover, this work raises an important, but as yet unanswered question: how can an agent optimally trade-off the efficiency of multi-task learning against multitasking capability? In this work, we: (a) show that this trade-off also arises in deep convolutional networks used to learn more complex tasks; (b) demonstrate that this trade-off can be managed by using single-task vs multitask training to control whether or not representations are shared; (c) propose and evaluate a meta-learning algorithm that can be used by a network to regulate its training and optimally manage the trade-off between multi-task learning and multitasking in an environment with unknown serialization costs.
\section{Background}

\begin{figure}
    \centering
    \includegraphics[width=0.4\textwidth]{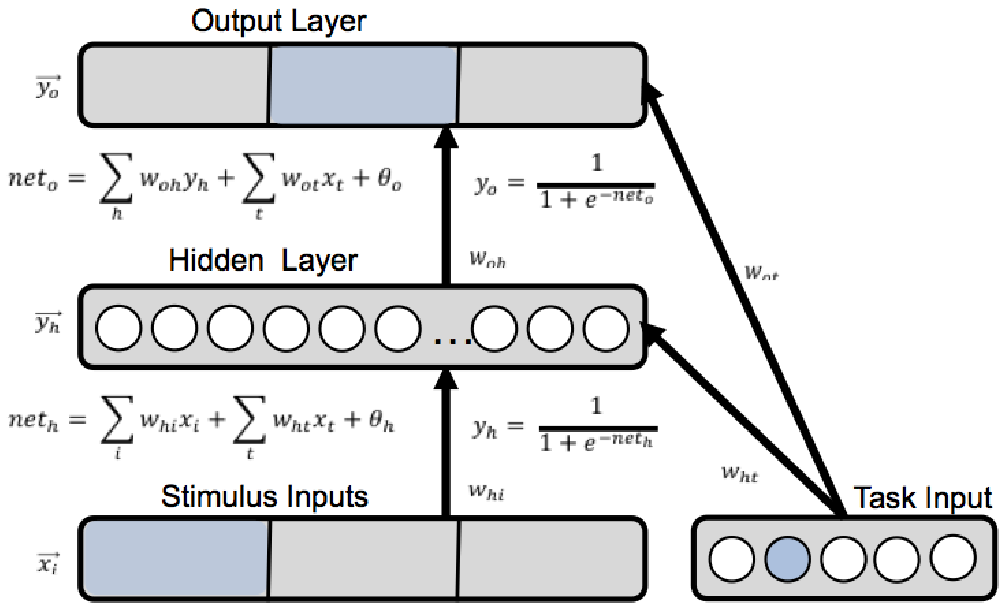}
    \caption{Neural network architecture from \citet{musslick2016controlled}.}
    \label{fig:example_NN}
\end{figure}

\begin{figure}
    \centering
    \includegraphics[width=0.45\textwidth]{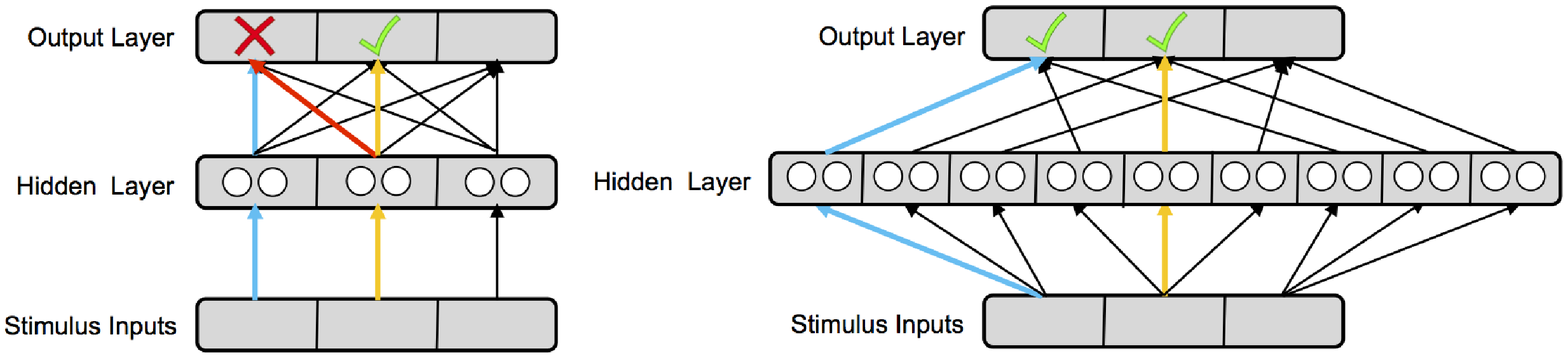}
    \caption{Network structure for minimal basis set (left) and tensor product (right) representations and the effects of multitasking in each. Red cross indicates error in execution of task because of interference whereas green check-mark indicates successful execution.}
    \label{fig:network_structure}
\end{figure}


\subsection{Definition of Tasks and Multitasking}

Consider an environment in which there are multiple stimulus input dimensions (e.g. corresponding to different sensory modalities) and multiple output dimensions (corresponding to different response modalities). Given an input dimension $I$ (e.g. an image) and an output dimension $O$ (e.g. object category) of responses, a task $T: I \to O$ represents a mapping between the two (e.g. mapping a set of images to a set of object categories), such that the mapping is independent of any other.
Thus, given $N$ different input dimensions and $K$ possible output dimensions, there is a total of $NK$ possible independent tasks that the network can learn to perform. Finally, multitasking refers to the simultaneous execution of multiple tasks, i.e. within one forward-pass from the inputs to the outputs of a network. Note that such multitasking differs from multi-task learning in that multitasking requires tasks to map different input dimensions to different output dimensions \cite{pashler1994dual} in a way that each is independent of the other, whereas typically in multi-task learning all tasks map the same input dimension to different output dimensions \cite{caruana1997multitask}. 

\subsection{Processing Single and Multiple Tasks Based on Task Projections}
\label{task-projection}
We focus on a network architecture that has been used extensively in previous work \cite{cohen1990control,botvinick2001conflict,musslick2017multitasking} (shown in Figure~\ref{fig:example_NN}). Here, in addition to the set of stimulus inputs, there is also an input dimension to indicate which task the network should perform. This task vector is projected to the hidden units and output units using learned weights. The hidden unit task projection biases the hidden layer to calculate a specific representation required for each task, whereas the output unit projection biases the outputs to only allow the output that is relevant for the task. The functional role of the task layer is inspired by the notion of cognitive control and attention in psychology and neuroscience, that is, the ability to flexibly guide information processing according to current task goals \cite{shiffrin1977controlled,posnerr,cohen1990control}. Assuming that the task representations used to specify different tasks are orthogonal to one another (e.g., using a one hot code for each), then multitasking can be specified by a superposition (sum) of the representations for the desired tasks in the task input layer. The weights learned for the projections from the task input units to units in the hidden layers, together with those learned within the rest of the network, co-determine what type of representation (shared or separate) the network uses.

\subsection{Minimal Basis Set vs Tensor Product Representations}

Previous work \citep{feng2014multitasking,musslick2016controlled,musslick2017multitasking} has established that, in the extreme, there are two ways that different tasks can be represented in the hidden layer of a two-layer  network. The first representational scheme is the \emph{minimal basis set} (shown on the left in Figure~\ref{fig:network_structure}), in which all tasks that rely on the same input encode the input in the same set of hidden representations. The second scheme is the \emph{tensor product} (shown on the right in Figure~\ref{fig:network_structure}), in which the input for each task is separately encoded in its own set of hidden representations. Thus, the minimal basis set maximally shares representations across tasks whereas the tensor product uses separate representations for each task.

These two representational schemes pose a fundamental trade-off. The minimal basis set provides a more efficient encoding of the inputs, and allows for faster learning of the tasks because of the sharing of information across tasks. However, it prohibits executing more than one task at a time (i.e. any multitasking). This is because, with the minimal basis set, attempting to execute two tasks concurrently causes the implicit execution of other tasks due to the representational sharing between tasks. In contrast, while the tensor product network scheme is less compact, multitasking is possible since each task is encoded separately in the network, so that cross-talk does not arise among them (see Figure \ref{fig:network_structure} for an example of multitasking and its effects in both types of networks). However, learning the tensor product representation takes longer since it cannot exploit the sharing of representations across tasks.

The type of representation learned by the network can be determined by the type of task-processing on which it is trained. Single-task training (referred to in the literature as multi-task training), involves training on tasks one at a time and generally induces shared representations. In contrast, multitask training involves training on multiple tasks concurrently and produces separate representations. This occurs because using shared representations when multitasking causes interference and thus error in task execution. In order to minimize this error and the cross-talk that is responsible for it, the network learns task projection weights that lead to separate representations for the tasks. In single-tasking training, there is no such pressure, as there is no potential for interference when executing one task at a time, and so the network can use shared representations. 
These effects have been established both theoretically and experimentally for shallow networks with one hidden-layer trained to perform simple synthetic tasks \citep{musslick2016controlled,musslick2017multitasking}. Below, we report results suggesting that they generalize to deep neural networks trained on more complex tasks.
\section{Meta-Learning for Optimal Agent}
The trade-off described above begs the following question: How can it be managed in an environment with unknown properties? That is, how does an agent decide whether to pursue single-task or multitask training in a new environment so that it can learn the tasks most efficiently while maximizing the rewards it receives?

Suppose an agent must learn how to optimize its performance on a given set of tasks in an environment over $\tau$ trials. At trial $t$, for each task, the agent receives a set of inputs $\mathbf{X} = \{\mathbf{x}_k\}_{k=1}^{K}$ and is expected to produce the correct labels $\mathbf{Y} = \{\mathbf{y}_k\}_{k=1}^{K}$ corresponding to the task. Assuming that each is a classification task, the agent's reward is its accuracy for the inputs, i.e. $R_t = \frac{1}{K} \sum_{k=1}^{K} \mathbbm{1}_{\hat{\mathbf{y}}_k = \mathbf{y}_k}$ where $\hat{\mathbf{y}}_k$ is the predicted label by the agent for each input $\mathbf{x}_k$. On each trial, the agent must perform all the tasks, and it can choose to do so either serially (i.e. by single-tasking) or simultaneously (i.e. by multitasking). After completion of the task and observation of the rewards, the agent also receives the correct labels $\mathbf{Y}$ for the tasks in order to train itself to improve task performance. Finally, assume that the agent's performance is measured across these trials through the entire course of learning and its goal is to maximize the sum of these rewards across all tasks.

To encode the time cost of single-tasking execution, we assume the environment has some unknown \emph{serialization} cost $c$ that determines the cost of performing tasks serially, i.e. one at a time. We assume that the reward for each task when done serially is $\frac{R_t}{1 + c}$ where $R_t$ is the reward as defined before. The serialization cost therefore discounts the reward in a multiplicative fashion for single-tasking. We assume that $0 \le c \le 1$ so that $c=0$ indicates there is no cost enforced for serial performance whereas $c=1$ indicates that the agent receives half the reward for all the tasks by performing them in sequence. Note that the training strategy the agent picks not only affects the immediate rewards it receives but also the future rewards, as it influences how effectively the agent learns the tasks to improve its performance in the future. Thus, depending on the serialization cost, the agent may receive lower reward for picking single-tasking but gains a benefit in learning speed that may or may not make up for it over the course of the entire learning episode. This question is at the heart of the trade-off the agent must navigate to make the optimal decision. We note that this is one simple but intuitive way to encode the cost of doing tasks serially but other mechanisms are possible.

\subsection{Approximate Bayesian Agent}
We assume that, on each trial, the agent has the choice between two training strategies to execute and learn the given tasks - by single-tasking or multitasking. The method we describe involves, on each trial, the agent modeling the reward dynamics under each training strategy for each task and picking the strategy that is predicted to give the highest discounted total future reward across all tasks. To model the reward progress under each strategy, we first define the reward function for each strategy, $f_{A,i}(t)$, which gives the reward for a task $i$ under strategy $A$ assuming strategy $A$ has been selected $t$ times. The reward function captures the effects of both the strategy's learning dynamics and unknown serialization cost (if it exists for the strategy). Here, $A \in \{S, M\}$ where $S$ represents the single-tasking strategy and $M$ represents the multitasking strategy.

We can use the reward function to get the reward for a task $i$ at trial $t'$ when selecting strategy $A$. Let $a_1, a_2, \ldots, a_{t' - 1}$ be the strategies picked at each trial until trial $t'$. Then,
\[
R^{(A,i)}_{t'} = f_{A,i}\left( \sum_{t=1}^{t'-1} \mathbbm{1}_{a_t = A} \right).
\]
is the reward for task $i$ at trial $t'$ assuming we pick strategy $A$.

Given the reward for each task, the agent can get the total discounted future reward for a strategy $A$ from trial $t'$ onward assuming we repeatedly select strategy $A$:
\[
R^{(A)}_{\ge t'} = \sum_{t=t'}^{\tau} \mu(t) \left( \sum_{i=1}^{N} R^{(A,i)}_{t} \right),
\]
where $\mu(t)$ is the temporal discounting function, $N$ is the total number of tasks, and $\tau$ is the total number of trials the agent has to maximize its reward on the tasks.

We now discuss how the agent maintains its estimate of each strategy's reward function for each task. The reward function is modeled as a sigmoidal function, the parameters of which are updated on each trial. Specifically, for a strategy $A$ and task $i$, using parameters $\theta_{A,i} = \{w_1, b_1, w_2, b_2 \}$, we model the reward function as $f_{A,i}(t) = \sigma(w_2 \cdot \sigma(w_1 \cdot t + b_1) + b_2)$.

We place a prior over the parameters $p(\theta_{A,i})$ and compute the posterior at each trial $t'$ over the parameters $p(\theta_{A,i} | D_{t'})$, where $D_{t'}$ is the observed rewards until trial $t'$ using strategy $A$ on task $i$. Because the exact posterior is difficult to compute, we calculate the approximate posterior $q(\theta_{A,i} | D_{t'})$ using variational inference \citep{wainwright2008graphical}. Specifically, we use Stein variational gradient descent (SVGD) \cite{liu2016stein}, which is a deterministic variational inference method that approximates the posterior using a set of particles that represent samples from the approximate posterior. The benefit of using SVGD is that it allows the number of particles used to be selected so as to increase the complexity of the approximate posterior, while ensuring that the time it takes to compute this approximation is practical. Because the posterior needs to be calculated repeatedly during training of the network, we found SVGD to offer the best properties - the approximate posterior is much quicker to compute than using MCMC techniques, while allowing for a complex approximation compared to using a simple Gaussian variational approximation to the posterior.

At each trial $t'$, the agent uses its estimate of the total discounted future reward for single-tasking and multitasking ($R^{(S)}_{\ge t'}$ and $R^{(M)}_{\ge t'}$, respectively) to decide which strategy to use. This can be thought of as a two-armed bandit problem, in which the agent needs to adequately explore and exploit to decide which arm, or strategy, is better. Choosing the single-tasking training regimen may give initial high reward (because of the learning speed benefit) but choosing multitasking may be the better long-term strategy because it does not suffer from any serialization cost. Thompson sampling \citep{thompson1933likelihood,chapelle2011empirical,gershman2018deconstructing} is an elegant solution to the explore-exploit problem, involving sampling from the posterior over the parameters and taking decisions greedily with respect to the sample. It provides initial exploration as the posterior variance is large at the start because of a lack of data and then turns to exploitation when the posterior is more confident due to seeing enough data. On each trial, we use Thompson sampling to pick between the training strategies by sampling from the approximate posterior over parameters for each strategy, calculating the total discounted future reward for each strategy according to the sampled parameters, and picking the strategy corresponding to the higher reward. Note that in practice we do not re-estimate the posterior in each trial (as one new reward will not change the posterior much) but instead do it periodically when enough new rewards have been observed.

\section{Related Work}
The most relevant work from the multi-task learning literature focuses on maximizing positive transfer across tasks while minimizing negative transfer. This list includes work on minimizing learning interference when doing multi-task training \citep{teh2017distral,rosenbaum2017routing} and reducing catastrophic interference when learning tasks one after another \citep{rusu2016progressive,kirkpatrick2017overcoming}.  However, to our knowledge, none of these works explicitly deal with the issue of how the type of representations a network uses affects whether it can execute tasks serially or in parallel.

Additionally, as mentioned previously, we build on previous work studying the trade-off of learning speed vs multitasking ability in artificial neural networks \citep{feng2014multitasking,musslick2016controlled,musslick2017multitasking}. Additionally, our meta-learning algorithm is similar to the one proposed by \citet{sagivefficiency}. However, we explicitly use the model's estimate of future rewards under each strategy to also decide how to train the network, whereas the meta-learner in \citep{sagivefficiency} was not applied to a neural network's learning dynamics. Instead, the actual learning curve for each strategy $A$ was defined according to pre-defined synthetic function. Our algorithm is thus applied in a much more complex setting in which estimation of each strategy's future rewards directly affects how the network chooses to be trained. Furthermore, our method is fully Bayesian in the sense that we utilize uncertainty in the parameter posterior distribution to control exploration vs exploitation via Thompson sampling. In \citep{sagivefficiency} logistic regression was combined with the $\epsilon$-greedy method to perform this trade-off, which requires hyper-parameters to control the degree of exploration. Lastly, we assume that the serialization cost is unknown and model its effects on the future reward of each strategy whereas \citep{sagivefficiency} makes the simplifying assumption that the cost is known. Modeling the effects of an unknown serialization cost on the reward makes the problem more difficult but is a necessary assumption when deploying agents that need to make such decisions in a new environment with unknown properties.

Lastly, previous work on bounded optimality \citep{russell1994provably,lieder2017strategy} is also relevant, as it is closely related to the idea of optimizing a series of computations given a processing cost as our proposed meta-learner does.
\section{Experiments}
In this section, we evaluate experimentally the aforementioned trade-off and proposed meta-learner model to resolve it. We first start by describing the task environment we use and the set of tasks we consider. We then describe the neural network architecture used, including the specific form of the task projection layer mentioned in section \ref{task-projection} that we use and how training occurs for single-tasking and multitasking. In sections \ref{tradeoff1} and \ref{tradeoff2}, we show through experiments explicitly how the trade-off arises in the task environment. Lastly, in section \ref{meta-learning}, we evaluate our proposed meta-learner's ability to navigate this trade-off in the environment given that there is an unknown serialization cost.

\subsection{Experimental Setup}
\begin{figure}
  \centering
  \includegraphics[width=3.8cm]{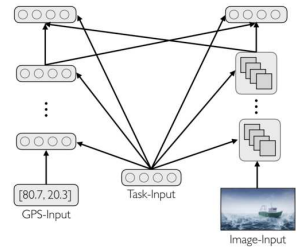}
  \caption{Neural network architecture used.}
  \label{fig:example_convNN}
\end{figure}

We create a synthetic task environment using AirSim \citep{shah2018airsim}, an open-source simulator for autonomous vehicles built on Unreal Engine\footnote{Code and data will be released in final version of paper.}. 
We assume a drone-agent that has two stimulus-inputs: (1) a GPS-input through which it can be given location-relevant information; (2) an image-input providing it visual information (e.g. from a camera). The agent also has two outputs: (1) a location-output designating a location in the input image; (2) an object-output designating the object the agent believes is present in the input. Based on the definition of a task as a mapping from one input to one output, this give us the following four tasks that the agent can perform:

\begin{enumerate}[itemsep=0mm, leftmargin=3\parindent]
\item[Task 1]  (GPS-localization): given a GPS location, output the position in the image of that location.

\item[Task 2] (GPS-classification): given a GPS location, output the type of object the agent expects to be in that area based on its experience.

\item[Task 3] (Image-localization): given a visual image,
output the location of the object in the image.

\item[Task 4] (Image-classification): given a visual image, output the type of object in the image.
\end{enumerate}

Using AirSim, we simulate an ocean-based environment with a set of different possible objects (such as whales, dolphins, orcas, and boats). We create training examples for the agent by randomizing the location of the agent within the environment, the type of object present in the visual input, the location and rotation of the object, and the GPS location provided to the agent. Thus, each training instance contains a set of randomized inputs and a label for each of the tasks with respect to the specific inputs. The agent can execute each task using either single-tasking (one after another) or multitasking (in which it can execute Tasks $1$ and $4$ together or Tasks $2$ and $3$ together). Note that in this setup, only $2$ tasks at most can be performed simultaneously as we will have conflicting outputs if we attempt to multitask more than $2$ tasks.

\subsection{Neural Network Architecture}
The GPS-input is processed using a single-layer neural network, whereas the image-input is processed using a multi-layer convolutional neural network. The encoded inputs are then mapped via fully-connected layers to each output. We allow the task input to modify each hidden, or convolutional, layer using a learned projection of the task input specific to each layer. This is related to the idea of cognitive control in psychology \citep{cohen1990control} but also to attention mechanisms used in machine learning \citep{hochreiter1997long}.

More formally, the task-specific projection for the $i^{\text{th}}$ layer $\mathbf{c}_i$ is computed using a matrix multiplication with learned task projection matrix $\mathbf{W}_{t,i}$ and task-input $\mathbf{x}_t$, followed by a sigmoid:
\[
\mathbf{c}_i = \sigma(\mathbf{W}_{t,i} \mathbf{x}_t - \beta),
\]
where $\beta$ is a positive constant. The subtraction by $\beta > 0$ means that task projections are by default ``off'' i.e. close to being $0$. For a fully-connected layer, the task projection $\mathbf{c}_i$ modifies the hidden units for the $i^{th}$ layer $\mathbf{h}_i$ through multiplicative gating to compute the hidden units $\mathbf{h}_{i+1}$:
\[
\mathbf{h}_{i+1} = g\left(
\left( \mathbf{W}_{h,i} \mathbf{h}_{i}  + \mathbf{b}_i \right) \odot \mathbf{c}_i \right),
\]
where $\mathbf{W}_{h,i}$ and $\mathbf{b}_i$ are the typical weight matrix and bias for the fully-connected layer, and $g$ is the non-linearity. For the hidden units, we let $g$ be the rectified linear activation function (ReLU) whereas for output units it is the identity function. Similarly, for a convolutional layer the feature maps $\mathbf{h}_{i+1}$ are computed from $\mathbf{h}_{i}$ as:
\[
\mathbf{h}_{i+1} = g\left(
\left(\mathbf{h}_{i} * \mathbf{W}_{h,i} + \mathbf{b}_i \right) \odot \mathbf{c}_i  \right),
\]
where $\mathbf{W}_{h,i}$ is now the convolutional kernel. Note that we use multiplicative biasing via the task projection whereas previous work \citep{musslick2016controlled,musslick2017multitasking} used additive biasing. We found multiplicative biasing to work better for settings in which the task projection matrix needs to be learned. A visual example of the network architecture is shown in Figure~\ref{fig:example_convNN}.

Training in this network occurs in the typical supervised way with some modifications. To train for a specific task, we feed in the stimulus-input and associated task-input, and train the network to produce the correct label at the output associated with the task. For outputs not associated with the task, we train the network to output some default value. In this work, we focus on classification-based tasks for simplicity, and so the network is trained via cross-entropy loss computed using the softmax over the network output logits and the true class label. To train the network on multitasking, we feed in the stimulus-input and the associated task-input (indicating which set of tasks to perform concurrently) and train the network on the sum of losses computed at the outputs associated with the set of tasks. Note that we consider the localization-based tasks as classification tasks by outputting a distribution over a set of pre-determined bounding boxes that partition the image space.

\begingroup
\makeatletter
\renewcommand{\p@subfigure}{}
\begin{figure}[t]
    \centering
    \begin{subfigure}[b]{0.5\textwidth}
      \centering
      \includegraphics[width=0.82\linewidth]{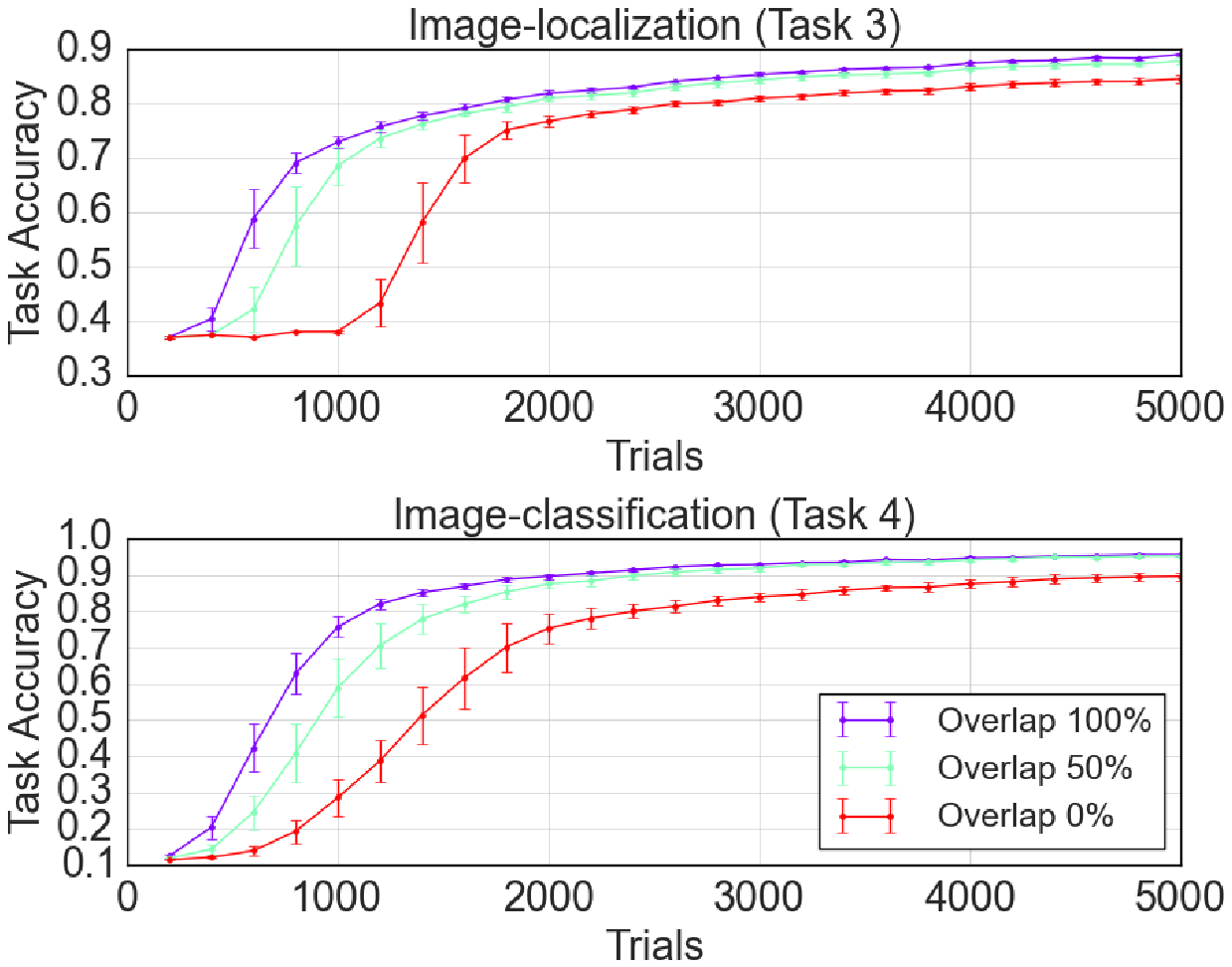}
      \caption{}
      \label{fig:overlap_learning_speed}
    \end{subfigure}
    \begin{subfigure}[b]{0.5\textwidth}
        \centering
        \begin{minipage}[b]{0.42\textwidth}
            \centering
            \includegraphics[width=\textwidth]{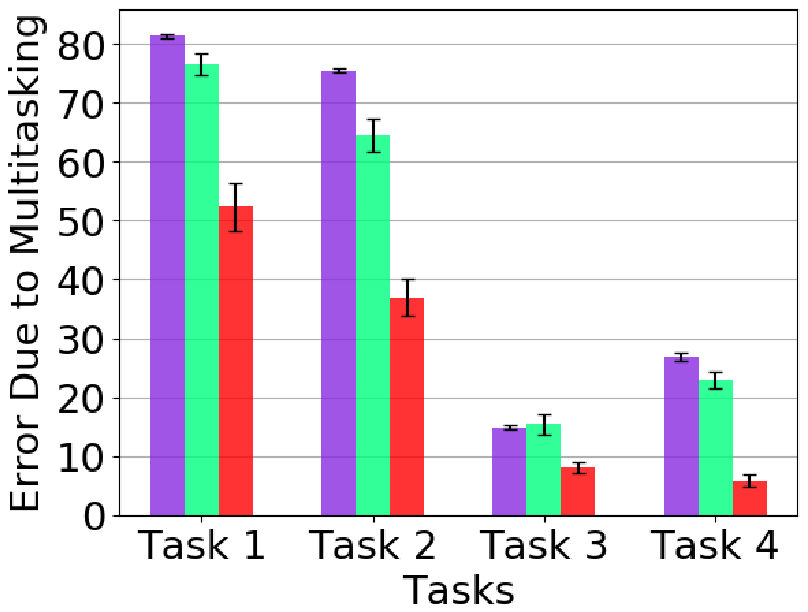}
            \caption{}
            \label{fig:overlap_error}
        \end{minipage}
        \begin{minipage}[b]{0.42\textwidth}
            \centering
            \includegraphics[width=\textwidth]{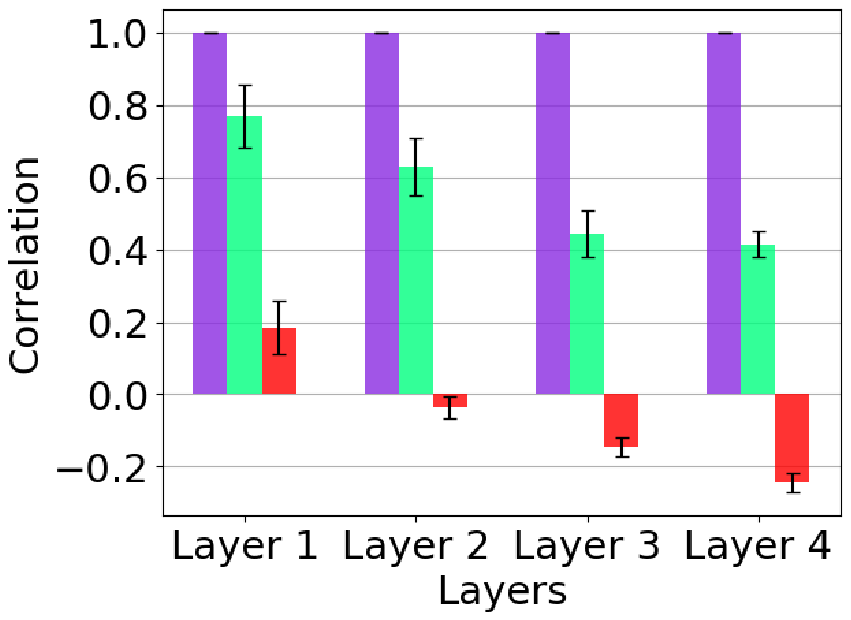}
            \caption{}
            \label{fig:overlap_corr}
        \end{minipage}
    \end{subfigure}
    \caption{Effect of varying representational overlap.  (\subref{fig:overlap_learning_speed}) Comparison of learning speed of the networks. (\subref{fig:overlap_error}) Comparison of the error in average task performance over all data when multitasking compared to single-tasking. (\subref{fig:overlap_corr}) Correlation of convolutional layer representations between Tasks $3$ and Tasks $4$ computed using the average representation for each layer across all the data. We show results for the tasks involving the convolutional network, as those are the more complex tasks we are interested in.}
    \label{fig:overlap}
\end{figure}
\endgroup

\subsection{Effect of Sharing Representations on Learning Speed and Multitasking Ability}
\label{tradeoff1}
First, we consider the effect of the degree of shared representations on learning speed and multitasking ability. We control the level of sharing in the representations used by the network by manipulating the task-associated weights $\mathbf{W}_{t,i}$, which implement, in effect, the task projection for each task. The more similar the task projections are for two tasks, the higher the level of sharing because more of the same hidden units are used for the two tasks. We vary $\mathbf{W}_{t,i}$ to manipulate what percent of hidden units overlap for the tasks. Thus, $100 \%$ overlap indicates that all hidden units are used by all tasks; $50 \%$ overlap indicates that $50 \%$ of the hidden units are shared between the tasks whereas the remaining $50 \%$ are split to be used independently for each task; and $0 \%$ overlap indicates that the tasks do not share any hidden units in a layer. Note that in this experiment, during training task-associated weights are frozen based on the initialization that results in the specific overlap percentage, but the weights in the remainder of the network are free to be learned. Based on previous work \citep{musslick2016controlled,musslick2017multitasking}, we measure the degree of sharing at a certain layer between two task representations by computing the correlation between the mean representation for the tasks, where the mean is computed by averaging the activity at the layer across all training examples for a given task.

The results of the experiment manipulating the level of overlap are shown in Figure~\ref{fig:overlap}. These show that as overlap is increased, sharing of representations across tasks increases (as evidenced by the increase in correlations), which is associated with an increase in the learning speed. However, this is associated with a degradation in the multitasking ability of the network, as a result of the increased interference caused by increased sharing of the representations. Note that the network with $0 \%$ overlap does not achieve error-free multitasking performance. This suggests that there is a residual amount of interference in the network induced by single-task training that cannot be attributed do the chosen manipulation i.e. overlap between task representations.

\subsection{Effect of Single-task vs Multitask Training}
\label{tradeoff2}

Having established that there is a trade-off in using shared representations in the deep neural network architecture described, we now focus on how different training regimens - using single-tasking vs multitasking - impact the representations used by the network and the network's learning speed. Previous work indicated that single-task training promotes shared representations and learning efficiency \cite{caruana1997multitask,musslick2017multitasking} whereas training a network to execute multiple tasks in parallel yields separated representations between tasks and improvements in multitasking \cite{MusslickCohen2019}.  
We compare different networks that vary on how much multitasking they are trained to do, from $0 \%$, in which the network is given only single-task training, to $90 \%$, in which the network is trained most of the time to do multitasking. Here, the task-associated weights $\mathbf{W}_{t,i}$ are initialized to be uniformly high across the tasks, meaning that the network is initially biased towards using shared representations, and all the weights (including task weights) are then learned based on the training regimen encountered by the network. We also conduct an experiment in which the network isn't as biased towards using shared representations by initializing smaller task-associated weights (see supplementary material). We note that the number of examples and the sequence of examples for each task are the same for both types of conditions (single-tasking or multitasking). The only difference is that in the case of single-task learning each task is learned independently using different forward and backward passes whereas in multitasking, multiple tasks can be processed together and thus learned together.

The results of this experiment (Figure~\ref{fig:multitask}) show that as the network is trained to do more multitasking, the learning speed of the network decreases and the correlation of the task representations also decreases. Because the network is initialized to use highly shared representations, we see that a multitasking training regimen clearly forces the network to move away from this initial starting point. The effect is stronger in the later layers, possibly because these layers may contribute more directly to the interference caused when multitasking.

\begingroup
\makeatletter
\renewcommand{\p@subfigure}{}
\begin{figure}[t]
    \centering
    \begin{subfigure}[b]{.5\textwidth}
      \centering
      \includegraphics[width=0.82\linewidth]{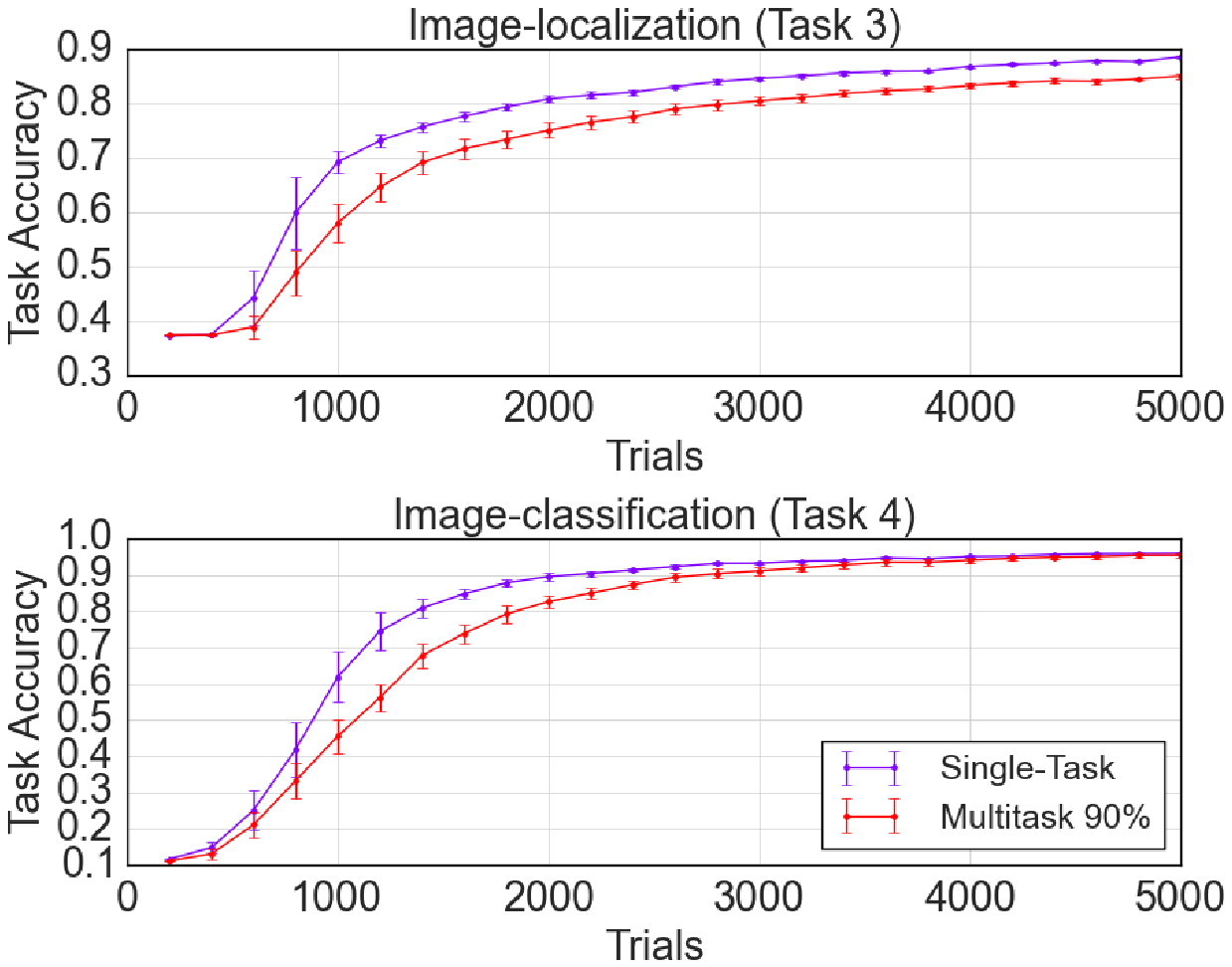}
      \caption{}
      \label{fig:multitask_learning_speed}
    \end{subfigure}
    \begin{subfigure}[b]{0.5\textwidth}
        \centering
        \begin{minipage}[b]{0.42\textwidth}
            \centering
            \includegraphics[width=\textwidth]{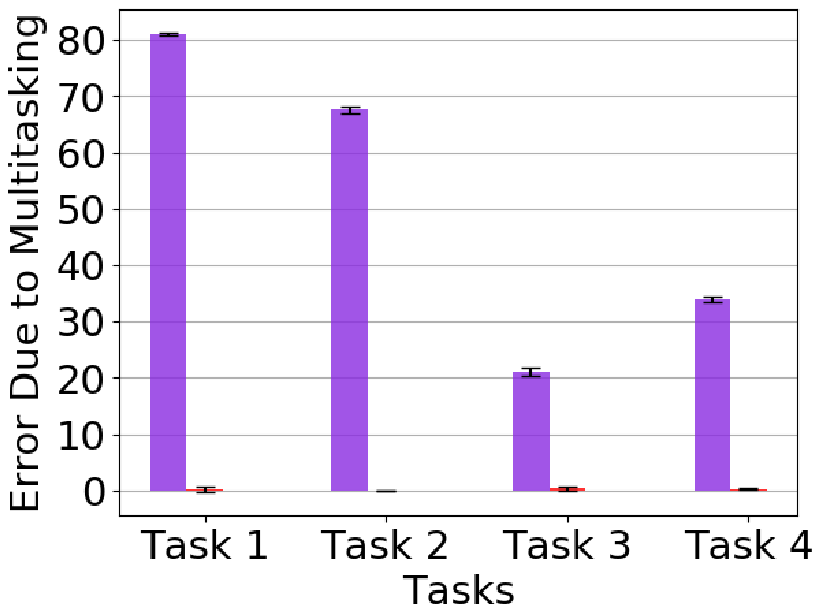}
            \caption{}
            \label{fig:multitask_error}
        \end{minipage}
        \begin{minipage}[b]{0.42\textwidth}
            \centering
            \includegraphics[width=\textwidth]{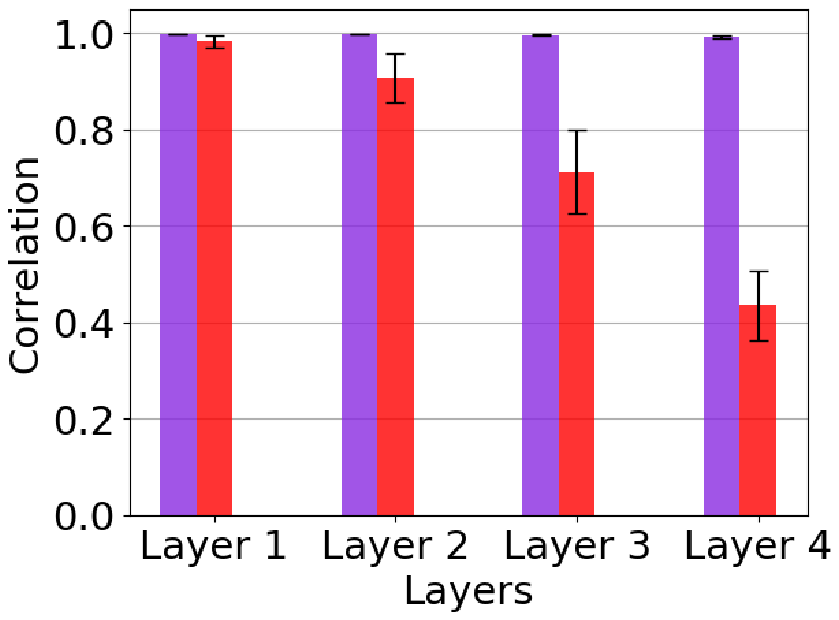}
            \caption{}
            \label{fig:multitask_corr}
        \end{minipage}
    \end{subfigure}
    \caption{Effect of single-task vs multitask training. (\subref{fig:multitask_learning_speed}) Comparison of learning speed of the networks. (\subref{fig:multitask_error}) Comparison of the error in average task performance over all data when multitasking compared to single-tasking (the lack of a bar indicates no error). (\subref{fig:multitask_corr}) Correlation of convolutional layer representations between Tasks $3$ and Tasks $4$ computed using the average representation for each layer across all the data. We again show results for the tasks involving the convolutional network.}
    \label{fig:multitask}
\end{figure}
\endgroup

\begin{figure}[t!]
    \centering
    \begin{subfigure}[b]{0.23\textwidth}
      \centering
      \includegraphics[width=\linewidth]{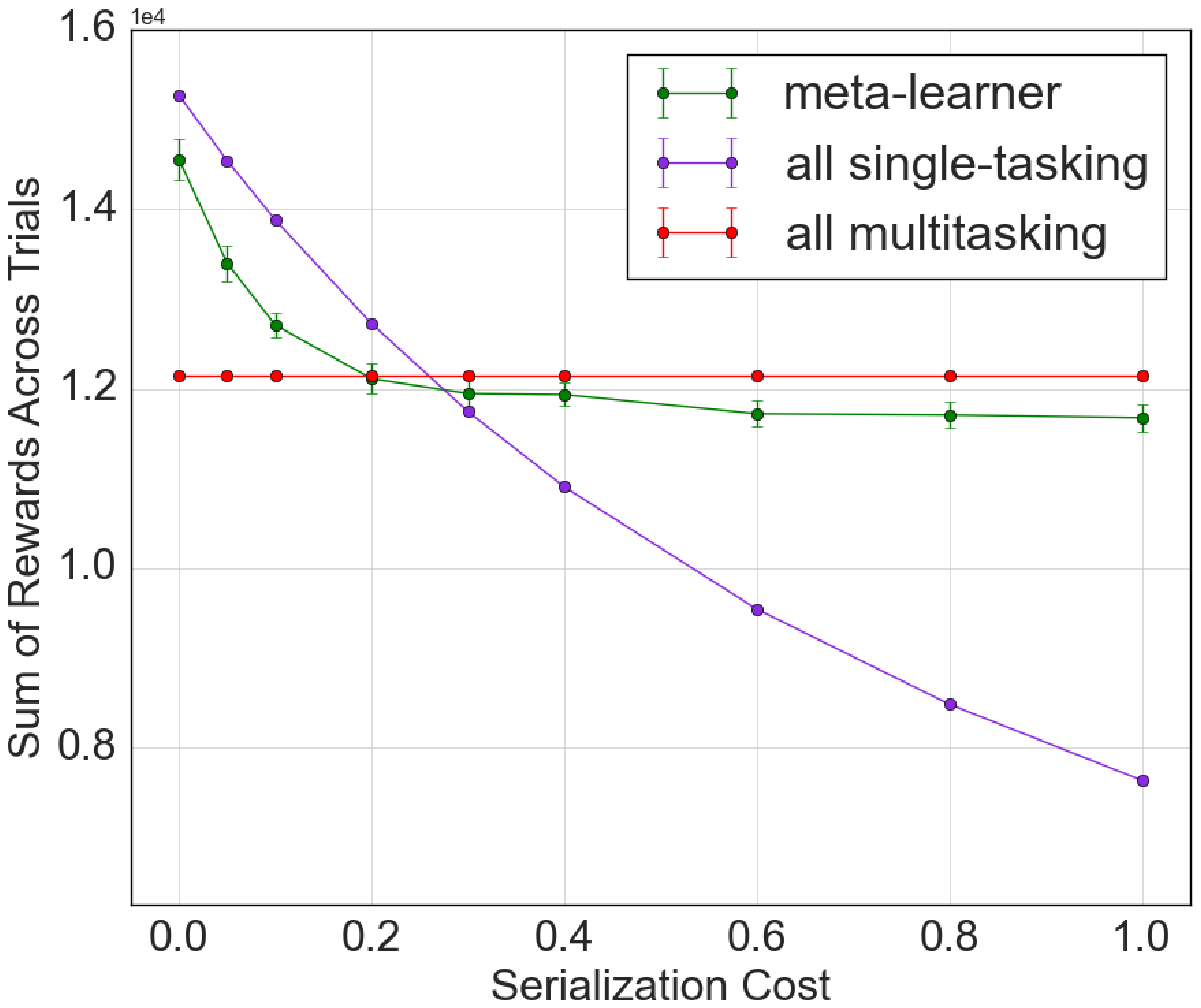}
      \caption{}
      \label{fig:meta_learner_eval1}
    \end{subfigure}
    \begin{subfigure}[b]{0.23\textwidth}
      \centering
      \includegraphics[width=\linewidth]{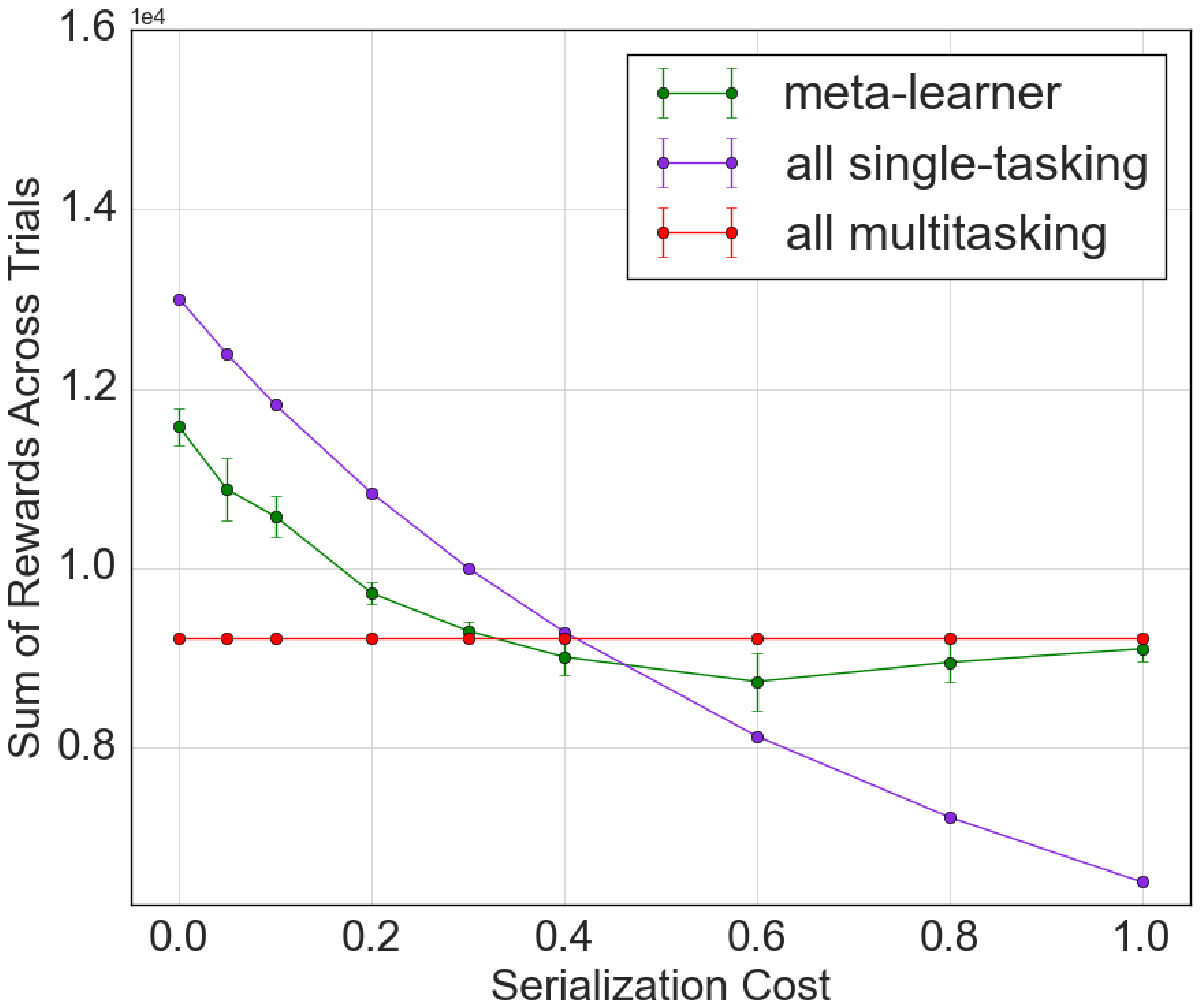}
      \caption{}
      \label{fig:meta_learner_eval2}
    \end{subfigure}
    \begin{subfigure}[b]{.23\textwidth}
      \centering
      \includegraphics[width=\linewidth]{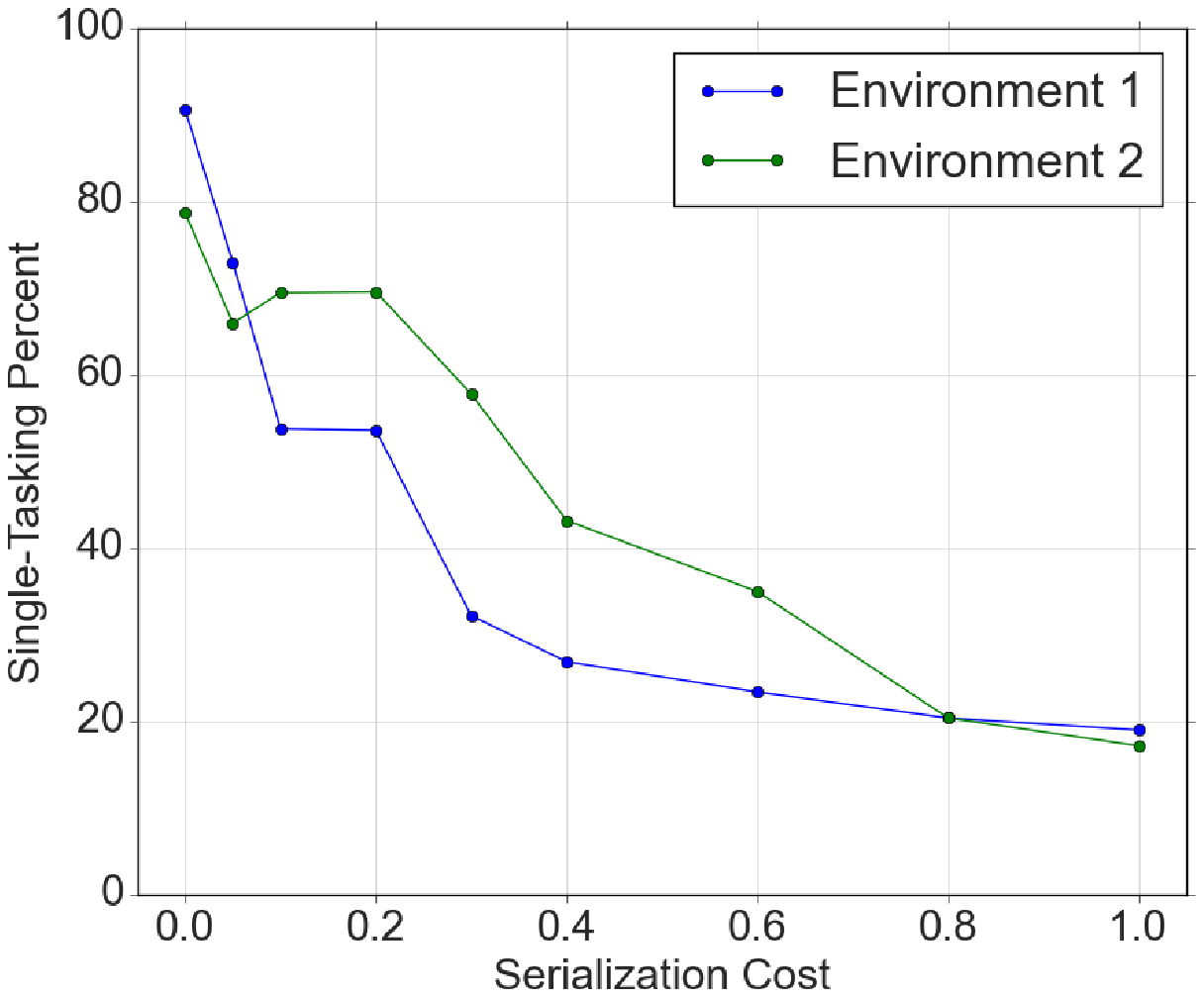}
      \caption{}
      \label{fig:single_percent}
    \end{subfigure}
    \caption{Evaluation of meta-learning algorithm. (\subref{fig:meta_learner_eval1}) Comparison of all methods on trade-off induced in original environment. (\subref{fig:meta_learner_eval2}) Comparison of all methods on trade-off induced in environment where noise is added to inputs. (\subref{fig:single_percent}) Percent of trials for which meta-learner picks to do single-tasking in both environments.}
    \label{fig:meta_learner_eval}
\end{figure}

\subsection{Meta-Learning}
\label{meta-learning}
Finally, having established the trade-off between single-task and multitask training, we evaluate the meta-learning algorithm to test its effectiveness in optimizing this trade-off. In order to test this in an environment with unknown serialization cost, we compare it with the extremes of always picking single-task or multitask training. We fix the total number of trials to be $\tau = 5000$ and evaluate each of the methods on varying serialization costs. For the meta-learner, we average the performances over $15$ different runs in order to account for the randomness involved in its sampling choices and measure its confidence interval. We fix the order in which data is presented for the tasks for all options when comparing them. Note that the meta-learner does not know the serialization cost and so has to model its effects as part of the received reward. We create two different environments to induce different trade-offs for rewards between single-tasking and multitasking. The first is a deterministic environment whereas in the second we add noise to the inputs. Adding noise to the inputs makes the tasks harder and seems to give bigger benefit to the minimal basis set (and single-task training). We hypothesize that this is the case because sharing information across tasks becomes more valuable when noisy information is provided for each task.

Figures \ref{fig:meta_learner_eval1} and \ref{fig:meta_learner_eval2} show that the meta-learning algorithm achieves a reward rate that closely approximates the one achieved by the strategy that yields the greatest reward for a given serialization cost. Additionally, note that in the extremes of the serialization cost, the meta-learner seems better at converging to the correct training strategy, while it achieves a lower reward when the optimal strategy is harder to assess. This difference is even clearer when we study the average percent of trials for which the meta-learner picks single-task training as a function of the serialization cost in Figure \ref{fig:single_percent}. 
We see that the meta-learning algorithm is well-behaved, in that as the serialization cost increases, the percent of trials in which it selects to do single-tasking smoothly decreases.
Additionally, at the points at which the optimal strategy is harder to determine, the meta-learner achieves reward closer to the worst strategy because it needs more time to sample each strategy before settling on one.
\section{Discussion}

In this work we study the trade-off between using shared vs separated representations in deep neural networks. We experimentally show that using shared representations leads to faster learning but at the cost of degraded multitasking performance\footnote{Note that limitations in multitasking due to shared representations may be bypassed by executing different single tasks across multiple copies of the trained network. However, this strategy appears inefficient as it requires a higher amount of memory and computation that scales with the number of tasks to be executed.}. We additionally propose and evaluate a meta-learning algorithm to decide which training strategy is best to use in an environment with unknown serialization cost. 

We believe simultaneous task execution as considered here could be important for real-world applications as it minimizes the number of forward passes needed to execute a set of tasks.  The cost of a forward pass is an important factor in embedded devices (in terms of both time and energy required) and scales badly as we consider larger task spaces and more complex networks. Thus, optimally managing the trade-off between learning speed vs multitasking could be crucial for maximizing efficiency in such situations.

A promising direction for future studies involves application of this meta-learner to more complex tasks. As we add more tasks, the potential for interference increases across tasks; however, as tasks become more difficult, the minimal basis set becomes more desirable, as there is even bigger benefit to sharing representations. Furthermore, in this more complicated setting, we would also like to expand our meta-learning algorithm to decide explicitly which set of tasks should be learned so that they can be executed in multitasking fashion and which set of tasks should only be executed one at a time. This requires a more complicated model, as we have to keep track of many possible strategies in order to see what will give the most reward in the future.

\bibliography{main}
\bibliographystyle{aaai}

\end{document}